%% file: main.tex
\documentclass[nohyperref]{article}
\usepackage[final]{neurips_2022/neurips_2022}
\usepackage{microtype}
\usepackage{standalone}
\usepackage{graphicx}
\usepackage{tabularx}
\usepackage{algorithm}
\usepackage[noend]{algpseudocode}
\usepackage{multirow}
\usepackage{comment}
\usepackage[linecolor=green!70!white, backgroundcolor=blue!20!white,
bordercolor=red, textcolor=black]{todonotes}
\usepackage{xcolor}
\usepackage{bm}
\usepackage{wrapfig}
\usepackage{tabularx}
\usepackage{caption}
\usepackage{appendix}
\usepackage{amsmath}
\usepackage{cancel}
\usepackage{float}
\usepackage{amssymb}
\usepackage{pifont}
\newcommand{\xmark}{\text{\ding{55}}}
\newcommand{\cmark}{\text{\ding{51}}}
\usepackage{subfigure}
\usepackage{booktabs} 
\usepackage{natbib}
\usepackage[colorlinks=true, linkcolor=red, citecolor=teal ]{hyperref}
\renewcommand{\th}{\theta}

\DeclareMathOperator*{\argmax}{argmax} 
\renewcommand{\L}{\mathcal{L}}
\newcommand{\KL}{\operatorname{KL}}

\newcommand{\divsep}{\,\|\,}               
\newcommand{\cond}{\,|\,}                  
\definecolor{darkblue} {rgb} {0.0 , 0.0 , 0.65}
\definecolor{darkred}  {rgb} {0.80, 0.0 , 0.0 }
\definecolor{darkgreen}{rgb} {0.0 , 0.50, 0.0 }
\definecolor{gray75}   {gray}{0.75}

\newcommand{\orange}[1]{{\color{orange} #1}}
\newcommand{\darkred}[1]{{\color{darkred} #1}}

\newcommand{\darkgreen}[1]{{\color{darkgreen} #1}}

\newcommand{\good}{\darkgreen{$\checkmark$}}

\newcommand{\bad}{\darkred{\xmark}}
\usepackage[capitalize,noabbrev]{cleveref}
\Crefname{equation}{Equation}{Equations}
\crefname{equation}{equation}{equations}

\title{Sparse Gaussian Process Hyperparameters: Optimize or Integrate?}

\author{%
Vidhi Lalchand \\
Department of Physics\\
University of Cambridge\\
\texttt{vr308@cam.ac.uk} \\
 \And
Wessel P. Bruinsma \\
Microsoft Research AI4Science \\
\texttt{wbruinsma@microsoft.com} \\
  \And
 David R. Burt \\
 LIDS \\
Massachusetts Institute of Technology \\
\texttt{dburt@mit.edu} \\
\And
 Carl E. Rasmussen \\
 Department of Engineering \\
University of Cambridge \\
\texttt{cer54@cam.ac.uk} \\
}

\begin{document}

\maketitle

\begin{abstract}

The kernel function and its hyperparameters are the central model selection choice in a Gaussian process \citep{rasmussen2005gaussian}.
Typically, the hyperparameters of the kernel are chosen by maximising the marginal likelihood, an approach known as \textit{Type-II maximum likelihood} (ML-II).
However, ML-II does not account for hyperparameter uncertainty, and it is well-known that this can lead to severely biased estimates and an underestimation of predictive uncertainty. While there are several works which employ a fully Bayesian characterisation of GPs, relatively few propose such approaches for the sparse GPs paradigm. In this work we propose an algorithm for sparse Gaussian process regression which leverages MCMC to sample from the hyperparameter posterior within the variational inducing point framework of \citep{titsias2009variational}. This work is closely related to \citet{hensman2015mcmc}, but side-steps the need to sample the inducing points, thereby significantly improving sampling efficiency in the Gaussian likelihood case. We compare this scheme against natural baselines in literature along with stochastic variational GPs (SVGPs) along with an extensive computational analysis.


\end{abstract}


\section{Introduction}

Gaussian processes (GPs) are a prominent class of models for supervised learning which can quantify uncertainty and incorporate inductive biases in function space via the kernel function. Hand-crafting a kernel function is a powerful way to incorporate prior knowledge. In many instances not all properties of a kernel function can be specified from prior knowledge alone, and parameters are chosen via ML-II. However, defining a complex kernel function with a large number of hyperparameters can make the marginal likelihood prone to multiple local optima and overfitting. Further, several local optima may correspond to priors that do not sensibly model the data. Weakly identified hyperparameters can manifest in flat ridges in the marginal likelihood surface\footnote{where different combinations of hyperparameters give very similar marginal likelihood values \vspace{2mm}} making gradient based optimisation extremely sensitive to starting values \citep{warnes1987problems}. Overall, the ML-II point estimates for the hyperparameters are subject to high variability and underestimate prediction uncertainty.

The problem of ridges in the marginal likelihood surface also does not necessarily go away as more observations are collected.
For example, if $f_1$ and $f_2$ are Brownian motions, $\sigma f_1(x / \ell)$ is equal in distribution to  $\sqrt{\alpha} \sigma f_2(x / \alpha \ell)$, which means observations do not provide any information about the product $\sigma \ell$. More generally, for a greater class of kernels, including the Mat\'ern--$1/2$ kernel, $\sigma f_1(x / \ell)$ is \emph{equivalent} to $\sqrt{\alpha} \sigma f_2(x / \alpha \ell)$, which means that is not possible to consistently estimate the product $\sigma \ell$ from data, no matter how many observations are collected in a fixed domain \citep[Chapter 6,][]{Stein:1999:Interpolation_of_Spatial_Data}. This implies that one cannot estimate the individual hyperparameters ($\sigma, \ell$) consistently. It also motivates why there can be benefits to estimating the hyperparameter posterior even in large data regimes, and ML-II may be insufficient. 
A more satisfactory treatment of hyperparameters involves placing a prior over the hyperparameters and performing Bayesian inference to compute a (hyper)posterior. For large datasets, this motivates using scalable GP inference (e.g.~sparse methods) in conjunction with Markov chain Monte Carlo (MCMC) for the hyperparametrs. 

\begin{figure}
    \centering
    \includegraphics[width=\textwidth]{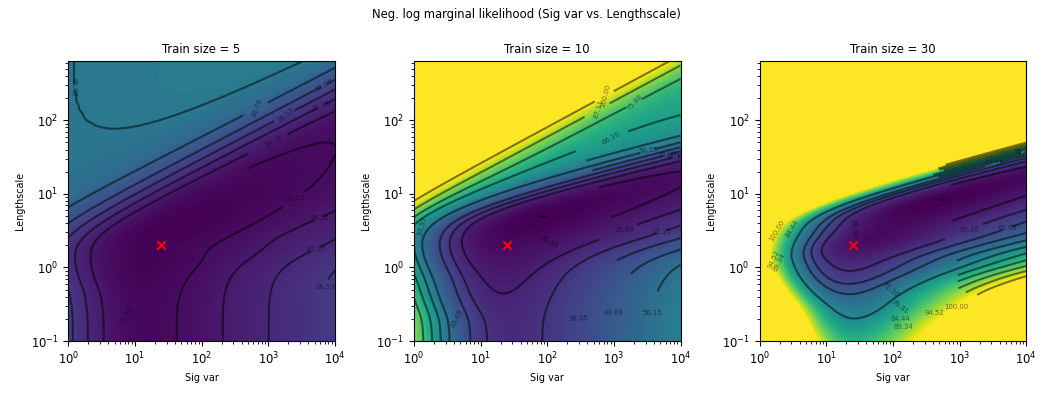}
    \caption{Negative log marginal likelihood surface as a function of two hyperparameters: $\sigma^{2}_{f}$ and $l$ for a squared exponential kernel and 1d function. The red cross indicates the true hyperparameters. The hyperparameters selected via gradien- based optimisation are sensitive to the initialisation 
    due to the long ridge of almost identical height at values of the hyperparameters not concordant with the ground truth.} 
    \label{fig:flat_ridges}
\end{figure}

The Bayesian treatment of weakly identified hyperparameters may also be fraught with difficulties. Gradient-based samplers like Hamiltonian Monte Carlo \citep{neal2011mcmc} and its variants have difficulty navigating regions of high curvature and flat ridges where the gradient offers no information for transition \citep{betancourt2017conceptual}. This leads to over-concentration of samples from the flat region. (Usually, this can at least partially be rectified with informative priors.)
These pathologies are also typical of other hierarchical models \citep{betancourt2015hamiltonian}. \Cref{fig:flat_ridges} shows that the GP marginal likelihood surface can manifest these pathologies. The evidence lower bound (ELBO) used in hyperparameter selection for variational sparse GPs relying on inducing points inherits similar, or even less favourable, characteristics to the exact marginal likelihood. As a result, for weakly informative priors, gradient based samplers are susceptible to getting \textit{stuck} at the boundary of these pathological regions hence biasing the sample estimates. The effective sample size metric, used for diagnosing mixing in MCMC, is indicative of this behaviour when directly observing the phase space of the target distribution, but is infeasible in high-dimensions. 

Historical justification for ML-II (also called the \textit{evidence} framework) comes from \citep{mackay1994hyperparameters} which highlighted several conditions for ML-II to yield reasonable estimates for the hyperparameters. Crucially, the evidence is unlikely to manifest multiple local optima for a model well-matched to the data and with a high signal-to-noise ratio. Transferring this insight to the Gaussian process paradigm we show how the evidence can have a significant tail or no well-defined maximum in settings with a low signal-to-noise ratio which arises with high aleatoric uncertainty or sparse data, frequently both. In these settings, a point estimate may not adequately summarise the hyperparameter posterior and the benefits of marginalisation stand out. The situation is only exacerbated in high-dimensions (i.e.~when there are many hyperparameters) where increasingly more volume of the posterior is captured in a thin shell making the density peak extremely unrepresentative of the posterior.
Unlike the likelihood of parameters, the marginal likelihood inherently contains a trade-off between the data-fit and complexity penalty term.
This is one of the main properties that makes the marginal likelihood objective a viable choice for model selection \citep{rasmussen2005gaussian}. For example, for the Gaussian process regression model,
\begin{equation}
    y_n = f(x_n) + \epsilon_n, \, \, \epsilon_n \sim \mathcal{N}(0, \sigma^2), \, f \sim \mathcal{GP}(0, k_{\theta})
\end{equation}
the marginal likelihood takes the form,
\begin{align}
\label{eqn:lml}
\log p(\bm{y}|\bm{\theta}) &= \log \int p(\bm{y}|f)p(f|\bm{\theta})df = c \smash{
    \overbrace{-\tfrac{1}{2}\bm{y}^{T}
              (K_{\theta} + \sigma^{2}I)^{-1}
              \bm{y}}^{\textrm{data fit term}}
  - \overbrace{\tfrac{1}{2}|K_{\theta} + \sigma^{2}I|}^{\textrm{complexity penalty}}
 }
\end{align}
where $c$ is a constant, $p(\bm{y}|f)$ denotes the data likelihood and $\bm{\theta}$ denotes kernel hyperparameters.
This trade-off is a well-established idea which  embodies the \textit{automatic Occam's razor} effect \citep{rasmussen2001occam} where models well-suited to the data are automatically selected just by using the marginal likelihood objective.

This may seem to contradict earlier claims regarding overfitting, but by shying away from dealing with the hyperparameter posterior we risk overfitting even with the marginal likelihood approach. In other words, the evidence framework subdues the overfitting effect induced by the canonical maximum likelihood approach which does not have any complexity penalty term. The parameters in the canonical approach are free to fit the data as well as possible making them prone to overfitting and poor generalisation. Overparameterized kernels based on neural networks like deep kernel models \citep{wilson2016deep} are well known to exacerbate overfitting \citep{ober2021promises}.

The main motivation for this work is to highlight that fully Bayesian schemes in sparse Gaussian process models are practically beneficial. While several works in the literature employ fully Bayesian scheme of integrating out the hyperparameters (see table \ref{references}), in this work we attempt to analyse them in an orthogonal direction, focusing on comparison with the evidence framework and other benchmarks by extending the main sparse variational formulation in the literature \citet{titsias2009variational}.
We present a generalised inference scheme for fully Bayesian GP regression and counteract some of the computational cost of sampling both inducing variables and hyperparameters by deriving a \textit{doubly collapsed} bound which selects the optimal distribution over the inducing points analytically and targets the kernel hyperparameters with HMC. 
\section{Related Work}
\begin{table*}[h]
    \centering
         \caption{Existing literature on fully Bayesian inference in GPs, sparse GPs and generic likelihoods.}
    \resizebox{\columnwidth}{!}{
    \begin{tabular}{c|l|c|l|l|l}
        \toprule
         Index & Reference & Sparse & Posterior ($\bm{f}$/$\bm{u}$) & Posterior ($\bm{\theta}$) & Methods  \\
         \midrule
         1. & \citet{murray2010slice} & $\xmark$ & sampling / NA & sampling & Slice Sampling \\
         2. & \citet{filippone2013comparative} & $\xmark$ & sampling / NA & sampling & MH + HMC + MA-LA \\
         3. & \citet{filippone2014pseudo} & $\xmark$ & Gaussian / NA & sampling & Deterministic + Pseudo-Marginal\\
         4. & \citet{hensman2015mcmc} & $\cmark$ & Gaussian / sampling & sampling & HMC \\
         5. & \citet{bui2018partitioned} & $\cmark$ & Gaussian / (sampling $\&$ VFE) & sampling / VFE &  MCMC \\
         6. & \citet{lalchand2020approximate} & $\xmark$ & Gaussian / NA & sampling / VFE & NUTS / VI \\
         7. & \citet{rossi2021sparse} & $\cmark$ & Gaussian / sampling & sampling & SG-HMC \\
         8. & \citet{simpson2020marginalised} & $\xmark$ & Gaussian / NA & sampling & NUTS / Nested Sampling \\
         9. & \textcolor{blue}{This work} & $\cmark$ & Gaussian / (sampling $\&$ VFE) & sampling & HMC / NUTS \\
         \bottomrule
    \end{tabular}}
    \label{references}
    \vspace{1mm}
\end{table*}
Fully Bayesian Gaussian processes have been used by several authors spawning several variants. In early accounts, \citet{neal1998}, \citet{williams1996gaussian} explore the integration over covariance hyperparameters using HMC in the regression setting. \citet{barber1997gaussian} extend this to the classification setting using HMC for sampling in the hyperparameter space and Laplace approximation for the integrand over function values. \citet{murray2010slice} and \citet{filippone2013comparative} focused on MCMC schemes to sample covariance hyperparameters in conjunction with latent function values, mainly mitigating the coupling effect through reparameterisation. \citet{hensman2015mcmc} considered joint sampling of inducing variables and hyperparameters from the optimal variational posterior distribution while \citep{bui2018partitioned} consider inference schemes for fully Bayesian sparse GPs in a streaming setting. More recently, \citet{rossi2021sparse} studied fully Bayesian sparse GPs using SG-HMC. \Citet{rossi2021sparse} modify the generative model by adding a prior over the inducing inputs, and peform inference using SG-HMC over the joint ($Z$, $\bm{u}$, $\theta$) space. 
We list the most recent works in Table \ref{references}. 

\section{Background}
\label{fbsgpr}
Let $f \sim \mathcal{GP}(0, k_{\bm{\theta}})$ be a Gaussian process prior with kernel function $k_{\bm{\theta}}$ depending on hyperparameters ${\bm{\theta}}$.
We are given noisy observations $\bm{y} = (y_{n})_{n=1}^{N} \subseteq \mathbb{R}$ of $\bm{f} = (f(\bm{x}_n))_{n=1}^N$ at input data $X = (\bm{x}_{n})_{n=1}^{N} \subseteq \mathbb{R}^D$.
We consider a Gaussian likelihood which factorises over the data, $p(\bm{y}|f) = \prod_{n=1}^{N}\mathcal{N}(y_{n}|f_{n}, \sigma^2)$. 
We wish to compute the posterior $p(f | \bm{y}, {\bm{\theta}})$.
In this section, we recapitulate the canonical inducing variable approximation of $p(f | \bm{y},{\bm{\theta}})$ by \citet{titsias2009variational} and its extension to a Bayesian treatment of the hyperparameters.

\subsection{Sparse variational inference in Gaussian processes}
Following \citet{titsias2009variational}, we consider a set of inducing variables $\bm{u} = \{f(\bm{z}_m)\}_{m=1}^{M} \subseteq \mathbb{R}$ at inducing inputs $Z = \{\bm{z}_{m}\}_{m=1}^{M}, \bm{z}_m \in \mathbb{R}^{D}$.
The complete generative model can then be factored as,
\begin{align}
    p(\bm{y}, f, \bm{u} | {\bm{\theta}}) = p(\bm{y} | f, {\bm{\theta}}) p(f | \bm{u}, {\bm{\theta}}) p(\bm{u} | {\bm{\theta}})
\end{align}

We approximate the posterior $p(f, \bm{u} | \bm{y}, {\bm{\theta}})$ with a variational distribution:
\begin{equation}
    p(f, \bm{u} | \bm{y}, {\bm{\theta}}) \approx q(f, \bm{u} | {\bm{\theta}}) = p(f | \bm{u}, {\bm{\theta}}) q(\bm{u})
\end{equation}

where $q(\bm{u})$ is chosen to minimise the Kullback--Leibler divergence $\operatorname{KL}(q(f | {\bm{\theta}}) \divsep p(f | \bm{y}, {\bm{\theta}}))$.
Minimising this KL divergence corresponds to maximising the evidence lower bound \citep{Matthews:2016:On_Sparse_Variational}, henceforth called the ELBO:
\begin{align}\label{eqn:elbo}
    \operatorname{ELBO}(q(\bm{u}), {\bm{\theta}})
    = \mathbb{E}_{q(f | {\bm{\theta}})}[&\log p(\bm{y} | f, {\bm{\theta}})] - \operatorname{KL}(q(\bm{u}|\bm{\theta}) \divsep p(\bm{u} | {\bm{\theta}}))
\end{align}

Because the ELBO still depends on $q(\bm{u})$, this bound is called \emph{uncollapsed}.
\citet{hensman2013gaussian,hensman2015scalable} let $q(\bm{u})$ be a Gaussian, which is optimal if the likelihood is Gaussian \citep{titsias2009variational}, approximate the expectation using Monte Carlo, and maximise the ELBO using stochastic optimisation.
On the other hand, if the likelihood is Gaussian, \citet{titsias2009variational} computes the optimal form for $q(\bm{u})$ directly:
\begin{align}
    q^*(\bm{u} | {\bm{\theta}}) = &\operatorname{argmax}_{q(\bm{u})}\,\operatorname{ELBO}(q(\bm{u}), {\bm{\theta}}) \propto p(\bm{u} | {\bm{\theta}}) \exp \mathbb{E}_{p(f | \bm{u}, {\bm{\theta}})}[\log p(\bm{y} | f, {\bm{\theta}})]
\end{align}
Plugging $q^*(\bm{u} | {\bm{\theta}})$ back into \cref{eqn:elbo}, the resulting bound is called \emph{collapsed}, because it now only depends on ${\bm{\theta}}$ and $Z$. The collapsed bound, denoted $\L_{\bm{\theta},Z}$, is the objective that \citet{titsias2009variational} proposes: 
\begin{align}\label{eqn:collapsed-elbo}
     \!\!\log p(\bm{y}|\bm{\theta}) &\geq \log \mathcal{N}(\bm{y}; \bm{0}, K_{nm}K_{mm}^{-1}K_{mn}\! +\! \sigma^{2}I) \!-\! \frac{1}{2\sigma^2}\textrm{Tr} (K_{nn}\!-\! K_{nm}K_{mm}^{-1}K_{mn}) =:  \mathcal{L}_{\bm{\theta}, Z},
\end{align}
where $K_{nn}$ is the prior covariance matrix of $\mathbf{f}$, $K_{mm}$ is the prior covariance matrix over $\mathbf{u}$ and $K_{nm}$ is crosss-covariance matrix formed by $\mathbf{f}$ and $\mathbf{u}$. Using the collapsed bound, approximate ML-II consists of finding,
\begin{equation}
      \bm{\theta}^* \in {\textstyle\argmax_{\bm{\theta},Z}}\, \mathcal{L}_{\bm{\theta},Z}.
      \vspace{2pt}
\end{equation}
Predictions at new functions values can be made in $O(M^2)$ after an initial cost of $O(NM^2)$. Under certain assumptions on the data generating process, even when $M \ll N$, the approximate posterior closely resembles the posterior, and \cref{eqn:collapsed-elbo} is a provably accurate approximation to \cref{eqn:lml} \citep{burt2020convergence}.

\subsection{Bayesian treatment of hyperparameters and sparse methods}
The extension of the sparse variational framework to a Bayesian treatment of the hyperparameters has been previously considered by \citet{hensman2015mcmc}.
Extend the generative model with a prior $p({\bm{\theta}})$ over the hyperparameters ${\bm{\theta}}$,
and let the variational approximation of the posterior $p(f, \bm{u}, {\bm{\theta}} | \bm{y})$ be $q(f, \bm{u}, {\bm{\theta}}) = p(f | \bm{u}, {\bm{\theta}}) q(\bm{u}, {\bm{\theta}})$.
The analogue of \cref{eqn:elbo} is 
\begin{align}\label{eqn:hyper-elbo}
    \operatorname{ELBO}&(q(\bm{u},{\bm{\theta}}))
    = \mathbb{E}_{q(f, {\bm{\theta}})}[\log p(\bm{y} | f, {\bm{\theta}})] - \mathbb{E}_{q({\bm{\theta}})}[\operatorname{KL}(q(\bm{u}) \divsep p(\bm{u} | {\bm{\theta}}))] - \operatorname{KL}(q({\bm{\theta}}) \divsep p({\bm{\theta}}))
\end{align}
and the optimal form for $q(\bm{u},{\bm{\theta}})$ can again be determined:
\begin{equation} \label{eq:optimal_q_joint}
    q^*(\bm{u},{\bm{\theta}})
    \propto p(\bm{u}, {\bm{\theta}}) \exp \mathbb{E}_{p(f | \bm{u}, {\bm{\theta}})}[\log p(\bm{y} | f, {\bm{\theta}})].
\end{equation}
The distribution $q^*(\bm{u}, \bm{\theta})$ does not have a closed form, and for general likelihoods, \citet{hensman2015mcmc} propose to approximate the expectation in \eqref{eq:optimal_q_joint} with quadrature and to sample from $q^*(\bm{u}, {\bm{\theta}})$ using HMC. While this approach is quite general, in the case of Gaussian regression, it vastly increases the dimensionality of the state space over which HMC must be run relative to HMC in GPR, since the $\bm{u}$ are sampled in addition to the $\bm{\theta}$. This increases the cost of the procedure, and impacts the success of the sampler. 

An alternative approach to approximately inferring hyperparameters is to assume a parametric form for $q(\bm{u}, \bm{\theta})$ and maximise \cref{eqn:hyper-elbo} with respect to the variational parameters. \Citet{bui2018partitioned} took such an approach, assuming that $q(\bm{u}, \bm{\theta}) = q(\bm{u})q(\bm{\theta})$, with both distributions Gaussian. Similar approaches have been applied to variational inference in state-space modelling, sometimes leveraging the optimal form of $q(\bm{u}|\bm{\theta})$ discussed earlier.

\begin{table*}
    \centering
    \caption{
        Comparison of a variety of approaches to approximating the posterior over hyperparameters in Gaussian process regression.
        Compares the quality of the posterior (\textsc{quality});
        the time complexity per iteration (\textsc{time/it.});
        the memory complexity per iteration (\textsc{mem./it.});
        the number of parameters and/or variables (\textsc{pars/vars});
        and whether the approach supports non-Gaussian likelihoods (\textsc{lik.}).
    }
    \newcommand{\spacer}{\hspace{0.5em}}
     \scalebox{0.75}{
    \begin{tabular}{lccccc}
        \toprule
        \textsc{approach} &
        \textsc{quality} &
        \textsc{time/it.} &
        \textsc{mem./it.} &
        \textsc{pars/vars} &
        \textsc{lik.} \\ \midrule
        Maximum a posteriori \citep{mackay1994hyperparameters} & \darkred{$\bm-$} & \darkred{$n^3$} & \darkred{$n^2$} & \darkgreen{$n_\theta$} & \bad \\
        \textsc{vi} \\
        \spacer Inducing points; non-collapsed \citep{titsias2014doubly} & \darkred{$\bm\pm$} & \darkgreen{$nm^2$} & \darkgreen{$m^2$} & \orange{$n^2_\theta + m^2$} & \good \\
        \spacer Inducing points; collapsed \citep{bui2018partitioned} & \darkred{$\bm\pm$} & \darkgreen{$nm^2$} & \darkgreen{$m^2$} & \darkgreen{$n^2_\theta$} & \bad \\
        \textsc{sampling} \\
        \spacer Exact with Gaussian lik.\citep{simpson2020marginalised} & \darkgreen{$\bm+$} & \darkred{$n^3$} & \darkred{$n^2$} & \darkgreen{$n_\theta$} & \bad \\
        \spacer Exact with non-Gaussian lik.\citep{Murray:2010:Slice_Sampling_Covariance_Hyperparameters_of} & \darkgreen{$\bm+$} & \darkred{$n^3$} & \darkred{$n^2$} & \darkred{$n_\theta + n$} & \good \\
        \spacer Inducing points; non-collapsed \citep{hensman2015mcmc}& \orange{$\bm \pm$} & \darkgreen{$m^3$} & \darkgreen{$m^2$} & \orange{$n_\theta + m$} & \good \\
        \spacer Inducing points; collapsed \textcolor{blue}{(ours)} & \orange{$\bm \pm$} & \darkgreen{$nm^2$} & \darkgreen{$m^2$} & \darkgreen{$n_\theta$} & \bad \\
        \bottomrule
    \end{tabular}}
    \label{tab:comparison_complexities}
    \vspace{2mm}
\end{table*}
The \textsc{quality} column in \cref{tab:comparison_complexities} indicates the ability of the method to faithfully represent the hyperparameter posterior. If VI is run to convergence, a potentially significant amount of error will be incurred by the Gaussian approximation to the non-Gaussian posterior over the hyperparameters (red). On the opposite extreme, if no sparsity assumption is made, MCMC over the hyperparameters without sparse approximations is asymptotically consistent (green). The inducing point approximations combined with MCMC lie somewhere in-between these methods (yellow).

\subsection{Making predictions}
The predictive posterior distribution for unknown test inputs $X^{*}$ integrates over the joint posterior,
\begin{align}
p(\bm{f}^{*}|\bm{y})&\approx \!\int \!p(\bm{f}^{*}| \bm{f}, \bm{u}, \bm{\theta})p(\bm{f} | \bm{u}, \bm{\theta})q(\bm{u}|\bm{\theta})q(\bm{\theta})d\bm{f}d\bm{u}d\bm{\theta},
\end{align}
where we have suppressed the conditioning over inputs $X, X^{*}$ for brevity. The inner integral simplifies to $\int p(\bm{f}^{*}| \bm{f}, \bm{u}, \bm{\theta})p(\bm{f} | \bm{u}, \bm{\theta})d\bm{f} = p(\bm{f}^{*}|\bm{u}, \bm{\theta})$. We discuss the predictive posterior in such models in section \ref{pred_pos}.

\section{Fully Bayesian SGPR with HMC: Doubly collapsed formulation}
\label{sec:sgpr_hmc}

In the previous section, we observed that a major drawback of the approach taken in \citet{hensman2015mcmc} is the need to sample $\bm{u}$, which for high-dimensional inputs or in cases where many inducing points are needed could introduce thousands of additional variables to sample. In this section, we leverage the optimal form of $q(\bm{u}|\bm{\theta})$ derived in \citet{titsias2009variational} to alleviate this sampling problem.

\subsection{Collapsing the evidence lower bound (again)}
\label{collapsing}
We first derive the lower bound for this formulation and provide pseudo-code for the algorithm in \cref{algmain}.
Following the usual derivation of the ELBO,
\begin{align}
\log p(\bm{y})
    &\ge \int q(\bm{\theta})  \log p(\bm{y} \cond \bm{\theta}) d\bm{\theta} - \KL(q(\bm{\th}) \divsep p(\bm{\th})) \\
         &\ge \int q(\bm{\th}) \mathcal{L}_{\bm{\theta}, Z}d\bm{\th} -  \KL(q(\bm{\th}) \divsep p(\bm{\th}))
         = \int q(\bm{\th})\log \dfrac{M_{\theta,Z}p(\bm{\theta})}{q(\bm{\theta})} d\bm{\th} =: \mathcal{L}^{*}_Z(q(\bm{\theta})), \label{eqn:doubly}
\end{align}
%
%
where
$\log p(\bm{y} \cond \bm{\theta}) \ge \L_{\bm{\theta},Z}$
with $\L_{\bm{\theta},Z}$ defined in \cref{eqn:collapsed-elbo}, and
where we assign
$M_{\bm{\theta}, Z} = e^{\L_{\bm{\theta},Z}}$. 

\subsubsection{Deriving \texorpdfstring{$q^{*}(\bm{\theta})$}{}}

We can interpret $\L^*_Z(q(\bm{\theta}))$ as a negative KL divergence as long as we account for a normalisation constant $C_Z = \int M_{\bm{\theta},Z}p(\bm{\theta}) d\bm{\theta}$ for the un-normalised numerator $M_{\bm{\theta},Z}p(\bm{\theta})$.
Hence, we can re-write $\L^*_Z(q(\bm{\theta}))$ as, 
\begin{equation} \label{eqn:doubly-2}
\L^*_Z(q(\bm{\theta})) =  \log C_Z - \KL(q(\bm{\th}) \divsep q^{*}(\bm{\theta}))
\end{equation}
where $q^{*}(\bm{\th}) = M_{\bm{\theta},Z}p(\bm{\theta})/ C_Z$.
By inspecting \cref{eqn:doubly-2}, we observe that the optimal variational distribution over $\bm{\theta}$ is given by $q^*(\bm{\theta})$.\footnote{KL divergence reaches its minimal value of zero when the two input probability distributions
are equal, and we seek to maximise $\mathcal{L}^{*}_Z(q(\bm{\theta}))$ which entails minimizing the KL.}
Crucially, by sampling directly from $q^{*}(\bm{\th})$ using MCMC we eliminate the need to sample the variables $\bm{u}$.
By evaluating $\L^*_Z$ at $q^{*}(\bm{\th})$, we find the \emph{doubly collapsed} ELBO $\L^{**}_Z := \L^*_Z(q^{*}(\bm{\th})) = \log C_Z$.
Although the value of $\L^{**}_Z$ is computationally intractable, given samples $(\bm{\th}_j)_{j=1}^J$ from $q^{*}(\bm{\th})$, gradients of $\L^{**}_Z$ with respect to $Z$ can be estimated using the stochastic estimate of the canonical ELBO \cref{eqn:collapsed-elbo}:
using the chain rule, 
\begin{equation} \label{eqn:wessel}
    \frac{\text{d}}{\text{d} Z}
    \mathcal{L}^{**}_{Z} 
    =
        \left.\frac{\partial}{\partial Z}
    \mathcal{L}^{*}_{Z}(q) 
    \right|_{q=q^*(\bm{\theta})}
        + 
        \left\langle
        \cancel{
            \left.\frac{\delta}{\delta q}
            \mathcal{L}^{*}_{Z}(q)\right|_{q=q^*(\bm{\theta})}
        }
        ,
        \frac{\partial}{\partial Z} q^*(\bm{\theta})
        \right\rangle
    \approx
     \frac1J \sum_{j=1}^J
        \frac{\partial}{\partial Z} 
            \mathcal{L}_{\bm{\theta}_j, Z}
\end{equation}
where $\frac{\delta}{\delta q}
            \mathcal{L}^{*}_{Z}(q)$
is the functional derivative of $\mathcal{L}^{*}_{Z}(q)$ with respect to $q$, which is zero at $q = q^*(\bm{\theta})$, because $q^*$ optimises $\mathcal{L}^{*}_{Z}$ (it is a critical point, so the derivative is zero). Further, the partial derivative of $\mathcal{L}^{*}_{Z}$ with respect to $Z$ concerns just the first term of the LHS of \cref{eqn:doubly} as the KL term $\KL(q(\bm{\theta}) \divsep p(\bm{\theta}))$ is independent of $Z$.

\subsection{Performing approximate inference}
We deploy HMC to (approximately) sample from the optimal variational posterior $q^{*}(\bm{\theta})$ along with optimising the inducing inputs $Z$ in a hybrid scheme. We alternate between the two steps allocating longer intervals for optimising $Z$ for every HMC sampling run for the hyperparameters. We note that this hybrid scheme is much more computationally efficient than sampling $\bm{u}$ and $\bm{\theta}$ jointly where one has to tackle the coupling between inducing variables and hyperparameters in joint space. Further, joint sampling is only feasible for moderate number of inducing variables while this scheme can scale to much larger datasets as the efficiency of sampling in the hyperparameter space is only dependent on the dimensionality of the hyperparameter space rather than the number of inducing variables. The entire inference scheme is summarized in \cref{algmain}. The warm-start strategy (of optimizing both  ($Z, \bm{\theta}$) jointly for a few gradient steps) is used to find a good region for the sampler to initialise $\bm{\theta}$.

\begin{algorithm}
\begin{algorithmic}[1]
\footnotesize
\caption{Fully Bayesian Sparse GPR with HMC} \label{algmain}
\State \textbf{Input:} ELBO objective $\mathcal{L}_{\theta, Z} := \mathcal{L}(\bm{\theta}, Z)$ (\cref{eqn:collapsed-elbo}), gradient based optimiser \texttt{optim()} \\
\Procedure{Warm-Start}{} 
\For{fixed number of steps}
  \State Gradient step: $Z, \bm{\theta} \longleftarrow \texttt{optim}(\mathcal{L}({\bm{\theta}, Z})$)
\EndFor \\
\Return{\textrm{initial values} $Z_{init}$ and $\bm{\theta}_{init}$}
\EndProcedure\\
\Procedure{Train}{}  
\State \texttt{ \#\# Initialisation protocol}
\State $\blacktriangleright$ Initialise $\mathcal{L}_{\theta, Z}$ at warm-start values $\mathcal{L}(\bm{\theta}_{init},Z_{init})$, lets call this $\mathcal{\hat{L}}$
  \State $\blacktriangleright$ Freeze kernel hyperparameters in the ELBO objective by setting \texttt{requires\_grad=False}. \\
  \While{not converged}
  
  \For {$t = 1 \ldots T$}  \texttt{ \#\# start of training loop} \\
  
    \State $\bullet$ Gradient step: $Z_{opt} \longleftarrow \texttt{optim}(\hat{\mathcal{L}})$ \texttt{\small{(\#\# \cref{eqn:wessel} shows the validity of taking the derivative of the stochastic ELBO)}}
    
  \If{$t \bmod L == 0$}
        \State\texttt{(\#\# For every L gradient steps)}\\
        
        \State  $\bullet$ Draw $J$ samples from the optimal hyperparameter variational
        distribution \State $\log q^{*}(\bm{\theta})$ $\propto \mathcal{L}({\bm{\theta}, Z_{opt}}) + \log p(\bm{\theta})$ keeping $Z_{opt}$ fixed. 
        $$  (\bm{\theta}_{j}, p_{j}) \xleftarrow{\text{HMC}}\mathcal{H}(\bm{\theta}, \bm{p}), \textrm{\texttt{(where $\mathcal{H}$ is the Hamiltonian)}}$$ 
        \State where $p$ denotes the zero-mean auxilliary momentum variable in phase space with 
        \State the same dimensionality as $\bm{\theta}$.
       \State $\bullet$  Compute stochastic ELBO $\hat{\mathcal{L}} = \dfrac{1}{J}\sum_{j=1}^{J}\mathcal{L}(\bm{\theta}_{j}, Z_{opt})$, where $\bm{\theta}_{j} \sim  q^{*}(\bm{\theta})$
    \EndIf 
 \EndFor 
 \EndWhile  \\
\Return{$Z_{opt}, \{\bm{\theta}\}_{j=1}^{J}$}
\EndProcedure
\end{algorithmic}
\end{algorithm}

\begin{figure*}[h]
    \includegraphics[scale=0.40]{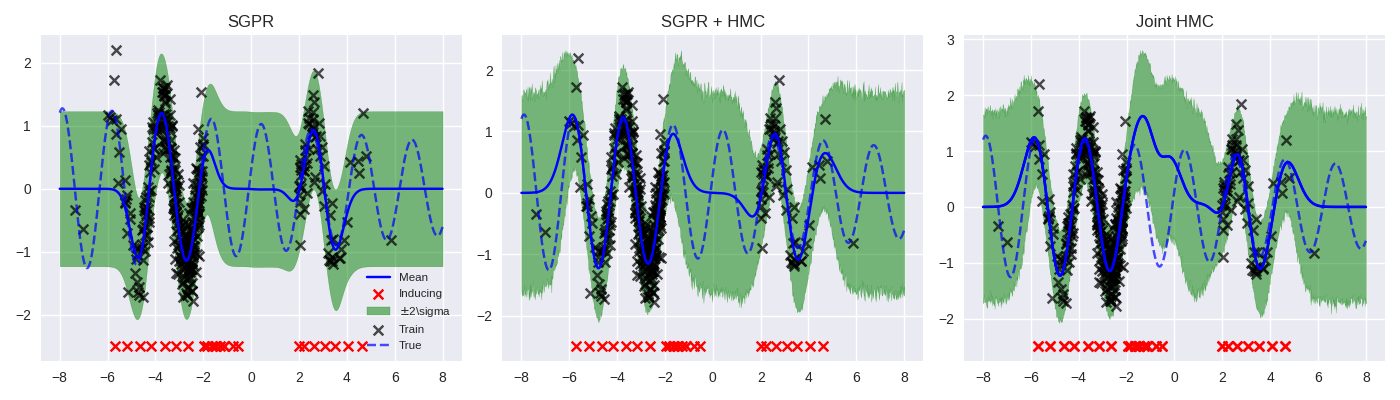}
    \includegraphics[scale=0.40]{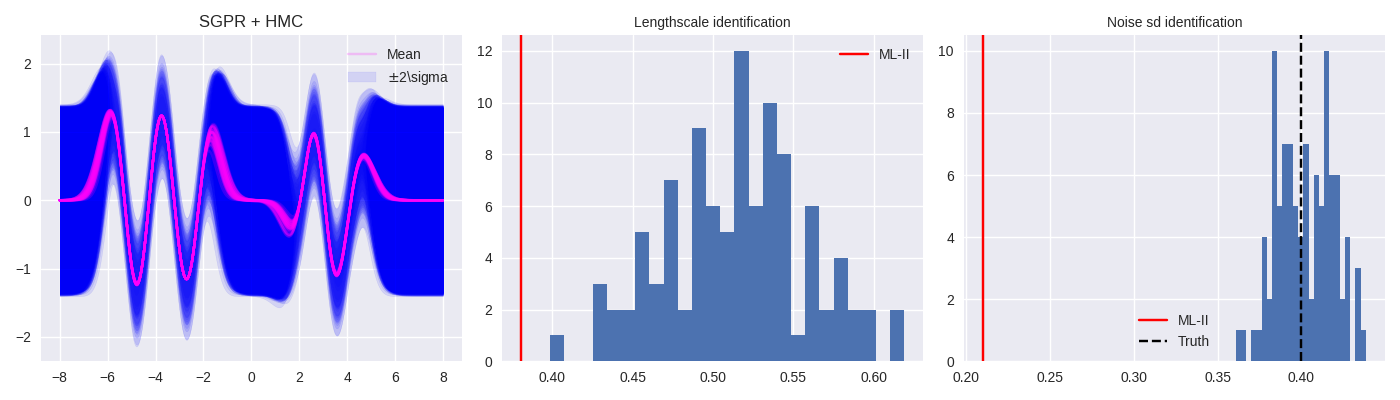}
    \caption{Top - 1d regression with Left: SGPR, Middle: SGPR $+$ HMC, Right: Joint HMC \citep{hensman2015mcmc}, Below - Left: Samples from the mixture posterior, Middle: Length-scale distribution under SGPR+HMC and ML-II. Note that the data is generated through a parameteric function and hence there is no ground truth lengthscale. Right: Noise std. deviation distribution from SGPR+HMC and ML-II.}
    \label{fig:syn}
\end{figure*}

\subsection{Predictive posterior}
\label{pred_pos}
It is ultimately the posterior predictive (PP) distribution that is of interest rather than point estimates or the hyperparameter posterior. The Bayesian sparse GP predictive posterior entails integrating out the posterior over inducing variables $\bm{u}$ and hyperparameters $\bm{\theta}$. Once we have performed inference, we can approximate this directly,
\begin{align}
    & p(\bm{f}^{*}|\bm{y}) = \int\int p(\bm{f}^{*}|\bm{u}, \bm{\theta})p(\bm{u},\bm{\theta}|\bm{y})d\bm{u}d\bm{\theta} = \int\int p(\bm{f}^{*}|\bm{u}, \bm{\theta})p(\bm{u}| \bm{\theta},\bm{y})p(\bm{\theta}|\bm{y})d\bm{u}d\bm{\theta} \\
    & \approx \int \int p(\bm{f}^{*}|\bm{u}, \bm{\theta})q^{*}(\bm{u}| \bm{\theta},\bm{y})p(\bm{\theta}|\bm{y})d\bm{u}d\bm{\theta} = \!\int\!\mathcal{N}(\bm{f^{*}}|A\bm{m}^{*}, K_{**} + A(S^{*}\! - \!K_{mm})A^{T})p(\bm{\theta}|\bm{y})d\bm{\theta} \nonumber 
    \label{pred}
\end{align}

where $q^{*}(\bm{u}|\bm{\theta}, \bm{y}) = \mathcal{N}(\bm{m}^{*}, S^{*})$ is the optimal Gaussian variational distribution and $A = K_{*m}K_{mm}^{-1}$ under the SGPR scheme (\cref{sec:sgpr_hmc}) is available in closed form with $\bm{m}^{*} = \sigma^{-2}(K_{mm} + \sigma^{-2}K_{mn}K_{nm})$ and $S^{*}=K_{mm}(K_{mm} + \sigma^{-2}K_{mn}K_{nm})^{-1}K_{mm}$. 
In either case, the internal integral with respect to the inducing variables is analytic and the outer integral can be estimated using the samples collected from HMC to perform Monte Carlo estimation. This  yields a mixture of Gaussians,
\begin{align}
p(\bm{f}^{*} | \bm{y}) \approx \frac{1}{J}{\textstyle\sum_{j=1}^{J}}\mathcal{N}(\bm{\mu}^{\theta_j}, \Sigma^{\theta_j}),&&  \bm{\theta_{j}} &\sim q^*(\bm{\theta}),
\\
\bm{\mu}^{\theta_j} = A^{(\theta_{j})}\bm{m}^{(\theta_{j})}, && 
\Sigma^{\theta_j} &= K_{**}^{\theta_{j}} \!+\! A^{(\theta_{j})}(S^{\theta_{j}} \!-\! K_{mm}^{(\theta_{j})})A^{{T}^{(\theta_{j})}} 
\end{align}
where $J$ samples are (approximately) drawn from $q^*(\bm{\theta})$ via HMC. The distribution inside the summation is the Gaussian posterior predictive distribution for fixed hyperparameters with identical mixing proportions.
The compute cost for the predictive posterior scales the sparse GPR cost linearly in the number of samples. The Monte Carlo approximation costs $\mathcal{O}(JNM^{2})$ for $M$ inducing points and $J$ hyperparameter samples. 

\section{Experiments}
\label{exp}
In the previous section, we discussed a hybrid scheme which leverages MCMC within the variational sparse inducing variable formulation leading to fully Bayesian sparse Gaussian processes. In the experiments we demonstrate the feasibility of this scheme relative to several benchmarks and assess regression performance on a 1-dimensional illustrative example and a range of other datasets. We also compare with exact GPs where the cost of using a gradient based sampler is prohibitive for even moderately large datasets requiring several inversions of the full covariance matrix. We show that using the doubly collapsed scheme proposed in this work is a much more attractive alternative for large datasets as compared to direct HMC in GPR, and sampling is more efficient than in the uncollapsed bound used in \citet{hensman2015mcmc}.

We henceforth refer to benchmark methods as follows: \textbf{SGPR + HMC} refers to Bayesian GPs with doubly collapsed variational inference with NUTS, as described in section \ref{fbsgpr} (compatible with Gaussian likelihoods); we benchmark this model against sparse GPs (\textbf{SGPR}) \citep{titsias2009variational} and Stochastic Variational GPs (\textbf{SVGP}) both using approximate ML-II \citep{hensman2015scalable}.
Additionally, we consider the sparse, joint sampling inference scheme proposed in \citep{hensman2015mcmc} which gives a natural benchmark. We call this model \textbf{JointHMC}. The \textbf{FBGP} method extends the Bayesian treatment to the inducing locations similar to \citet{rossi2021sparse}; we use NUTS to sample from the joint posterior over $(Z, \bm{\theta})$. It is not a direct comparison to \citet{rossi2021sparse} as the latter explores free-form sampling of $\bm{u}$ along with $(Z, \bm{\theta})$ while we work with the collapsed bound incorporating the optimal Gaussian variational distribution $q^{*}(\bm{u})$.

We also present analysis where we fix inducing point locations at a random subset of the training data (as opposed to interleaving as per \cref{algmain}) and only learn hyperparameters using NUTS. We provide several details about the experimental set-up in the supplementary. 

\begin{figure}[t!]
    \includegraphics[ clip,width=\columnwidth]{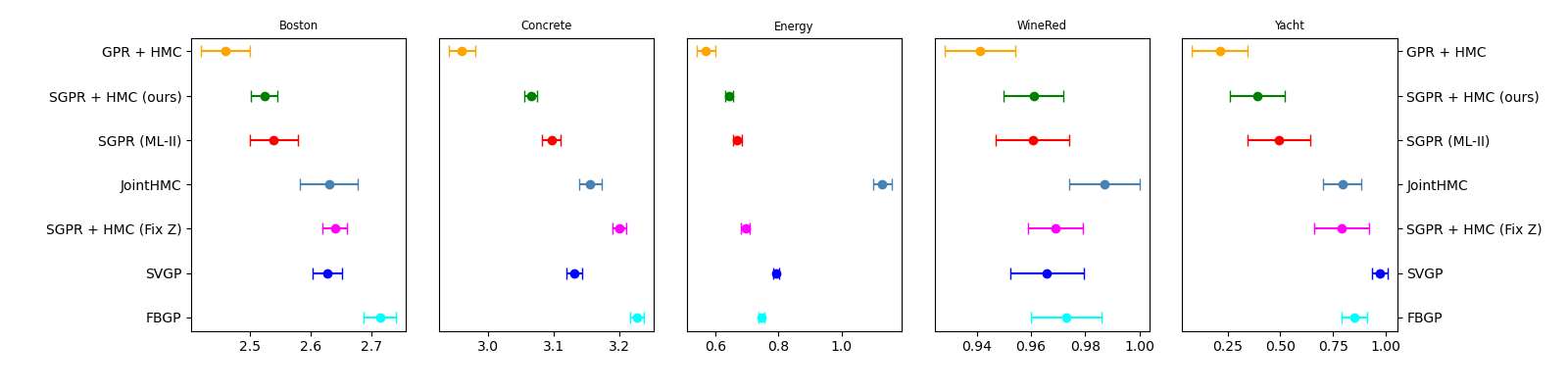}
    \caption{Negative test log likelihoods with standard error of mean across 10 splits with 80$\%$ of the data reserved for training. Our method is SGPR + HMC.}
    \vspace{5mm}
    \label{fig:nlpd}
\end{figure}

\subsection{One dimensional synthetic data}

\begin{wraptable}{r}{0.5\textwidth}
\caption{Prediction performance in 1D synthetic regression across SGPR, SGPR + HMC and JointHMC methods with identical number of inducing points and train/test split. }\label{table:1d-metrics}
\small
\centering
\begin{tabular}{l|c|c|c}
\toprule
Method & SGPR & SGPR + HMC & JointHMC \\
\midrule
RMSE & 0.580 & \textbf{0.537} & 0.682 \\
NLPD & 0.214 & \textbf{0.065} & 0.74 \\
\bottomrule
\end{tabular}
\end{wraptable} 

We sample noisy observations from $f(x) = \sin(3x) + 0.3\cos(\pi x)$ with the constraint $(x < -2)$ and $(x > 2)$. \cref{fig:syn} shows the results for SGPR along with the fully Bayesian schemes. We keep data split and noise identical across the three models to facilitate a comparison. While there is significant data to identify the hyperparameters we notice that the models mainly differ in their extrapolation abilities away from the training data. SGPR with ML-II overfits to the training data and recovers a low lengthscale, low noise solution while the SGPR + HMC scheme recovers a more moderate fit and performs significantly better in terms of RMSE and NLPD on unseen data \cref{table:1d-metrics}. We note that the JointHMC scheme which samples both $(\bm{u}, \bm{\theta})$ overfits in the central missing data region. We use $M=25$ inducing locations across all methods which are optimised according to the protocol of each method and recover a very similar spatial distribution. 
\subsection{UCI regression benchmarks}
\input{tables/uci_regression}
We compare our approach across methods on 5 standard small to medium-sized UCI benchmark datasets. Following common practice, we use a 20$\%$ randomly selected held out test-set \citep{rossi2021sparse, havasi2018inference} and scale the inputs and outputs to zero mean and unit standard deviation within the training set (we restore the output scaling for evaluation) \citep{salimbeni2017doubly}. While we could use any kernel, we choose the RBF-ARD kernel with a lengthscale for each dimension. For consistency we initialise all the inducing locations ($Z$) identically across the methods, i.e.\ by using the same random subset of training data split. We note that adapting the inducing locations brings serious gains in prediction performance versus keeping them fixed (\cref{fig:nlpd}). Further, the JointHMC scheme underperforms SGPR (with ML-II) and SGPR + HMC. This is not surprising given that the JointHMC bound \cref{eq:optimal_q_joint} does not incorporate the optimal setting for $q(\bm{u})$ and was originally motivated by the need for a fully Bayesian scheme for generalised likelihoods. The method SGPR + HMC significantly improves upon JointHMC in the specific Gaussian likelihood case. 

\textbf{Sensitivity to $M$}: We benchmark SGPR+HMC and JointHMC on the Elevator dataset ($N=16599, D=18$) which demands a larger $M$. SGPR+HMC outperforms JointHMC for this dataset across different $M$ but the advantage is more pronounced at smaller $M$. Further, our method took 1248 vs. 2109 wall clock sec. for the joint scheme for the same number of hyperparameter samples and 500 inducing points.
\subsection{Ablation study}
In order to understand the efficacy of \cref{algmain} we conduct an ablation study where we perform inference in the same manner, but keeping inducing locations fixed. Algorithmically, this implies that we don't need to compute the stochastic ELBO $\mathcal{\hat{L}}$ and just conduct a single sampling run for the hyperparameters. The results across 10 splits are summarised in \cref{tab:anlpds}.
\begin{figure}
\centering
{\vspace{-1mm}
\includegraphics[scale=0.6]{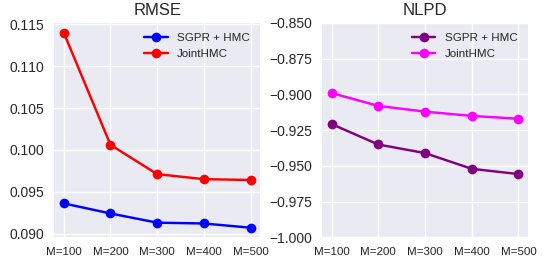}}
\includegraphics[scale=0.45]{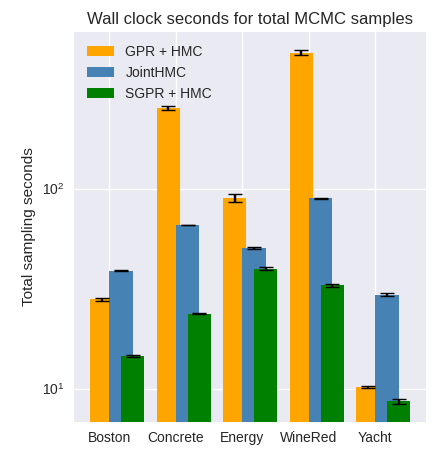}
 \caption{Left: Test RMSE and NLPD for a range of inducing points for the Elevator dataset. Right: Sampling performance measured in terms of the time it took to draw the combined set of samples during the training phase (excluding tuning) as defined by the python standard library \textit{time.perf\_counter (wall time)}. We use the \texttt{pymc3} \texttt{pm.NUTS} sampler for GPR + HMC and SGPR + HMC, and  \texttt{tfp.mcmc.HamiltonianMonteCarlo} for JointHMC \citep{matthews2017gpflow}. All experiments were conducted on an Intel Core i7-10700 CPU @ 2.90GHz x 16.}
 \label{fig:secs}
     \vspace{3mm}
\end{figure}
\input{tables/ablation_nlpds}

\subsection{Runtimes} While it is possible to train exact GPR with HMC for datasets of this size (in terms of $N$) it is important to look at the trade-off in terms of compute cost. In \cref{fig:secs} we record the average number of wall clock seconds to draw 500 samples under each method. The cost of sampling is  $\mathcal{O}(N)$ for SGPR + HMC but $\mathcal{O}(N^{3})$ for Exact GPR + HMC. JointHMC deals with a higher dimensional phase space $(\bm{u}, \bm{\theta})$ hence requires more tuning. We don't include tuning time for a fair comparison. For further context we report the total training run-time for our scheme alongside ML-II, GPR + HMC and FBGP in \cref{tab:trainrun}. The hybrid scheme we propose is significantly cheaper to canonical alternatives with virtually no degradation in predictive performance.
\begin{table}[H]
\caption{Wall clock seconds (this counts all the CPU time, including worker processes in BLAS and OpenMP as defined by the python standard library \texttt{time} for a single training split).}
\vspace{5pt}
\label{tab:trainrun}
\resizebox{\columnwidth}{!}{
\begin{tabular}{l|l|l|l|l|l}
\toprule
Dataset      & Boston       & Concrete       & Energy          & WineRed         & Yacht \\
\midrule
SGPR (ML-II) & 22.17 (0.21) & 33.06 (0.07)   & 30.36 (0.09)    & 39.96 (0.87)    & 20.41 (0.22)  \\
SGPR + HMC (\textcolor{blue}{ours}) & 29.47 (0.34) & 53.85 (2.36)   & 61.60 (1.47)    & 60.65 (0.63)    & 24.50 (0.40)  \\
GPR + HMC    & 78.05 (2.36) & 977.40 (13.82) & 326.18 (15.87) & 1426.25 (39.49) & 31.71 (0.59)  \\
FBGP         & 72.63 (8.29) & 156.31 (4.30)  & 259.81 (11.58)  & 175.45 (13.14)  & 101.92 (2.27) \\
\bottomrule
\end{tabular}}
\end{table}

\section{Discussion}
The evidence framework continues to be the pre-dominant method for training Gaussian processes since their inception into modern machine learning \citep{rasmussen2005gaussian}. While the marginal likelihood is a compelling model selection objective as it offers an inherent trade-off between data-fit and complexity, it is susceptible to overfitting and other pathologies leading to biased inference. This work builds on existing methods that combine sparse Gaussian process regression based on inducing variables with Bayesian hyperparameter inference.

Bayesian hyperparameter inference in GPs is however intractable and one has to consider balancing between objectives of computational cost, prediction accuracy and robustness of uncertainty intervals. While in straightforward conditions the fully Bayesian approach might be counter-productive, most real-world applications of GPs rely on engineering sophisticated hand-crafted kernels involving many hyperparameters where there risk of overfitting is pronounced and further, harder to detect. A more robust solution is to incorporate prediction intervals that reflect these uncertainties in the model choice. Studying full Bayesian inference in more sophisticated GP models like deep \citep{damianou2013deep}, warped \citep{snelson2004warped} and convolutional GPs \citep{van2017convolutional} will offer greater insight to this question and is an imminent direction of future work. 
\section*{Acknowledgements}
This research was conducted while WPB and DRB were students at the University of Cambridge.
During that time, WPB was supported by the Engineering and Physical Research Council (studentship number 10436152). VL acknowledges funding from the Qualcomm Innovation Fellowship (Europe). 
\bibliographystyle{plainnat}
\bibliography{ref}

\section*{Checklist}

\begin{enumerate}

\item For all authors...
\begin{enumerate}
  \item Do the main claims made in the abstract and introduction accurately reflect the paper's contributions and scope?
    \answerYes{We provide extensive background to justify.}
  \item Did you describe the limitations of your work?
    \answerYes{We include a brief discussion in the supplementary.}
  \item Did you discuss any potential negative societal impacts of your work?
    \answerNo{This work is largely methodological but we touch upon this briefly in the limitations section in the supplementary.}
  \item Have you read the ethics review guidelines and ensured that your paper conforms to them?
    \answerYes{}
\end{enumerate}

\item If you are including theoretical results...
\begin{enumerate}
  \item Did you state the full set of assumptions of all theoretical results?
    \answerYes{We provide derivations with discuss assumptions in section \cref{sec:sgpr_hmc}}
        \item Did you include complete proofs of all theoretical results?
    \answerNA{}
\end{enumerate}

\item If you ran experiments...
\begin{enumerate}
  \item Did you include the code, data, and instructions needed to reproduce the main experimental results (either in the supplemental material or as a URL)?
    \answerYes{We include assumptions and other experimental details like the prior hyperparameters, learning rate and data splits in the supplementary.}
  \item Did you specify all the training details (e.g., data splits, hyperparameters, how they were chosen)?
    \answerYes{We include these details along with a pymc3 code snippet for reproducibility.}
        \item Did you report error bars (e.g., with respect to the random seed after running experiments multiple times)?
    \answerYes{We report std error of the mean across 10 splits for most experiments.}
        \item Did you include the total amount of compute and the type of resources used (e.g., type of GPUs, internal cluster, or cloud provider)?
    \answerYes{Since we mainly use small to moderate sized datasets we conducted most experiments on the CPU.}
\end{enumerate}

\item If you are using existing assets (e.g., code, data, models) or curating/releasing new assets...
\begin{enumerate}
  \item If your work uses existing assets, did you cite the creators?
    \answerNA{}
  \item Did you mention the license of the assets?
    \answerNA{}
  \item Did you include any new assets either in the supplemental material or as a URL?
    \answerNA{}
  \item Did you discuss whether and how consent was obtained from people whose data you're using/curating?
    \answerNA{}
  \item Did you discuss whether the data you are using/curating contains personally identifiable information or offensive content?
    \answerNA{}
\end{enumerate}

\item If you used crowdsourcing or conducted research with human subjects...
\begin{enumerate}
  \item Did you include the full text of instructions given to participants and screenshots, if applicable?
    \answerNA{}
  \item Did you describe any potential participant risks, with links to Institutional Review Board (IRB) approvals, if applicable?
    \answerNA{}
  \item Did you include the estimated hourly wage paid to participants and the total amount spent on participant compensation?
    \answerNA{}
\end{enumerate}
\end{enumerate}

\appendix
\begin{appendices}
\section{Experimental Set-up}

For methods SGPR, SGPR + HMC, JointHMC and Ablation experiment we use the Adam \citep{kingma2014adam} optimizer with a learning rate set at 0.01 (we didn't extensively tune for learning rates and 0.01 seemed to give a reasonable performance). We do maintain consistency over the data splits and initialisation values for the inducing locations and hyperparameters across all the methods. Further, all the sparse models use $M=100$ inducing variables to aid in run-time analysis. All the hyperparameters are initialised at the \texttt{gpytorch} default of $\log(2)$ and inducing locations at a random subset of the training data split.

\textbf{SGPR + HMC:} We place individual priors over the set of hyperparameters $\{\{l_{d}\}_{d=1}^{D}, \sigma_{f},  \sigma_{n}\}$ shown in the code block below. During the warm-up phase we optimize both the inducing locations and hyperparamters. We use $J=100$ samples to construct the stochastic ELBO for the first sampling window along with 500 steps of tuning, thereafter just 10 samples are used every 50 gradient steps. At the end of training we again draw $J=100$ samples. The intermediate sampling windows do not require elaborate tuning as we persist good initial step-size values from the penultimate chains. Despite this we do expend a few tuning steps in each sampling window as it improved the overall performance of the sampler. The inducing locations are kept fixed during sampling and are only optimized through the stochastic ELBO.

\textbf{JointHMC:} As recommended by the authors we use a warm-up phase of 100 gradient steps to optimize inducing locations. Subsequent training happens through the HMC sampler which targets the joint variables $(\bm{v}, \bm{\theta})$ (where $\bm{v}$ is a whitened representation of $\bm{u}$) with a target acceptance rate of 0.8, path length (number of leapfrog steps) to 10 and an initial step-size of 0.01 with an adaptation rate of 0.1. We use \texttt{tfd.Gamma(2.0, 1.0)} for each inidividual kernel hyperparamter.

\subsection{Software $\&$ Code}

The software for all the methods is largely written in \texttt{gpytorch} \citep{gardner2018gpytorch}. For sampling we resort to the auto-tuning NUTS sampler in \texttt{pymc3} \citep{salvatier2016probabilistic}. The JointHMC model uses the SGPMC class from \texttt{gpflow} \citep{GPflow2020multioutput}. The source code for all the models and experiments is attached with the supplementary. 

The code-snippet below shows the straight-forward \texttt{pymc3} sampling loop which is triggered at pre-specified intervals. 
\footnotesize{
\begin{verbatim}
with pm.Model() as model_pymc3:
   
    ls = pm.Gamma("ls", alpha=2, beta=1, shape=(input_dim,))
    sig_f = pm.HalfCauchy("sig_f", beta=1)

    cov = sig_f ** 2 * pm.gp.cov.ExpQuad(input_dim, ls=ls)
    gp = pm.gp.MarginalSparse(cov_func=cov, approx="VFE")
    sig_n = pm.HalfCauchy("sig_n", beta=1)
    
    # Z_opt is the intermediate inducing points from the optimisation stage
    y_ = gp.marginal_likelihood("y", X=self.train_x.numpy(), Xu=Z_opt, \\ 
                                                y=self.train_y.numpy(), noise=sig_n)
                                                                                    
    if sampler_params is not None:
        step = pm.NUTS(step_scale = sampler_params['step_scale'])
    else:
        step = pm.NUTS()
    trace = pm.sample(n_samples, tune=tune, chains=1, step=step, \\
                                                return_inferencedata=False)  
return trace
\end{verbatim}}

\section{Further Analysis}

\subsection{Comparison with Deep GPs and Neural Network Benchmarks}

We additionally compare the performance of our algorithm to  2, 3 and 4 layer deep GPs (DGP
2–4), each with 100 inducing points and point estimation for the hyperparameters \citep{damianou2013deep}, \citep{salimbeni2017doubly}. We also compare to a two-layer Bayesian neural network with ReLu activations, 50 hidden units, with inference by probabilistic backpropagation (PBP). The results were taken from \citet{salimbeni2017doubly} and \citet{hernandez2015probabilistic} respectively and follow a very similar data processing scheme for the datasets. We learn the inducing locations $Z$ through optimisation but keep the number of inducing points fixed across all methods.

\begin{figure}[h]
\hspace{-5mm}
    \includegraphics[width=1.09\textwidth]{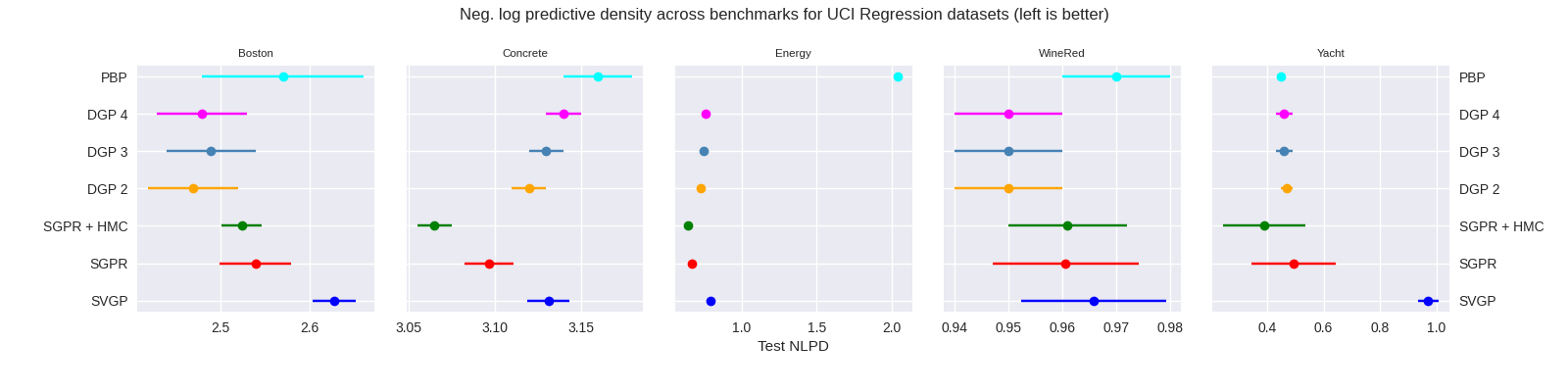}
    \caption{Negative test log likelihoods with standard error of mean with 80$\%$ of the data reserved for training. Our method is SGPR + HMC.}
    \label{fig:nlpd2}
\end{figure}
The negative test log-likelihood results are shown in \ref{fig:nlpd2}. The test log-likelihoods outperform the non-Bayesian counterparts and in most cases perform as well if not better than a multi-layer deep GP with a significantly higher computational cost and intractabilities. Further, the variability across splits is much lower for the HMC method versus SGPR. 

\subsection{NUTS Sampling Summary}

In the tables below we include the summary statistics of the NUTS sampler for split 4 for each dataset for the SGPR + HMC model. The statistics were computed based on the trace of the final sampling window.
The columns hdi$\_$3\% and hdi$\_$97\% calculate the highest posterior density interval based on marginal posteriors. $\texttt{ess} = \dfrac{MN}{1 + 2\sum_{t=1}^T \hat{\rho}_t}$ computes effective sample size where $M$ is the number of chains, $N$ is the number of samples in each chain and $\rho_{t}$ denotes auto-correlation at lag $t$. For the results reported below $N=100$ and $M=1$. Each chain was run with 500 warm-up iterations for the sampler to adapt to an optimal step-size. 
\texttt{ess\_{bulk}} refers to the effective sample size based on the rank normalized draws and is a useful indicator of sampling efficiency. \texttt{ess\_{tail}} computes the minimum of the effective sample sizes of the 3\% and 97\% quantiles \citep{vehtari2021rank}.

\subsection{Boston}

\begin{table}[H]
\centering
\begin{tabular}{l|l|l|l|l|l|l|l|l}
hyper      & mean  & sd    & hdi\_3\% & hdi\_97\% & mcse\_mean & mcse\_sd & ess\_bulk & ess\_tail \\
\hline
ls{[}0{]}  & 2.415 & 0.627 & 1.464    & 3.685     & 0.073      & 0.055    & 97.0      & 58.0      \\
ls{[}1{]}  & 7.236 & 1.641 & 4.814    & 10.737    & 0.137      & 0.105    & 165.0     & 71.0      \\
ls{[}2{]}  & 5.751 & 1.56  & 2.903    & 8.099     & 0.169      & 0.13     & 104.0     & 69.0      \\
ls{[}3{]}  & 8.42  & 1.71  & 6.041    & 11.732    & 0.154      & 0.109    & 120.0     & 112.0     \\
ls{[}4{]}  & 3.711 & 1.308 & 1.708    & 6.042     & 0.141      & 0.1      & 89.0      & 113.0     \\
ls{[}5{]}  & 3.366 & 0.402 & 2.731    & 4.155     & 0.035      & 0.025    & 141.0     & 78.0      \\
ls{[}6{]}  & 5.594 & 1.235 & 3.363    & 7.957     & 0.117      & 0.086    & 109.0     & 60.0      \\
ls{[}7{]}  & 3.078 & 0.926 & 1.524    & 4.89      & 0.092      & 0.065    & 97.0      & 77.0      \\
ls{[}8{]}  & 6.53  & 1.515 & 4.034    & 9.074     & 0.148      & 0.106    & 104.0     & 62.0      \\
ls{[}9{]}  & 2.416 & 0.64  & 1.425    & 3.8       & 0.054      & 0.039    & 146.0     & 78.0      \\
ls{[}10{]} & 5.388 & 1.505 & 2.815    & 8.137     & 0.143      & 0.106    & 108.0     & 74.0      \\
ls{[}11{]} & 6.239 & 2.198 & 3.307    & 10.616    & 0.267      & 0.203    & 75.0      & 77.0      \\
ls{[}12{]} & 1.808 & 0.316 & 1.248    & 2.34      & 0.036      & 0.026    & 85.0      & 77.0      \\
sig\_f     & 1.067 & 0.15  & 0.796    & 1.333     & 0.017      & 0.013    & 75.0      & 59.0      \\
sig\_n     & 0.277 & 0.01  & 0.261    & 0.293     & 0.001      & 0.001    & 180.0     & 87.0     
\end{tabular}
\end{table}

\subsection{Yacht}

\begin{table}[H]
\centering
\begin{tabular}{l|l|l|l|l|l|l|l|l}
hyper     & mean   & sd    & hdi\_3\% & hdi\_97\% & mcse\_mean & mcse\_sd & ess\_bulk & ess\_tail \\
\hline
ls{[}0{]} & 7.486  & 1.042 & 5.522    & 9.25      & 0.049      & 0.037    & 500.0     & 343.0     \\
ls{[}1{]} & 10.361 & 1.496 & 7.657    & 13.045    & 0.067      & 0.048    & 498.0     & 320.0     \\
ls{[}2{]} & 15.365 & 2.588 & 10.592   & 19.864    & 0.103      & 0.075    & 643.0     & 397.0     \\
ls{[}3{]} & 12.464 & 2.12  & 8.818    & 16.494    & 0.073      & 0.053    & 836.0     & 499.0     \\
ls{[}4{]} & 15.372 & 2.543 & 10.137   & 19.902    & 0.092      & 0.066    & 740.0     & 307.0     \\
ls{[}5{]} & 1.368  & 0.088 & 1.215    & 1.536     & 0.005      & 0.003    & 360.0     & 427.0     \\
sig\_f    & 2.334  & 0.341 & 1.765    & 3.051     & 0.019      & 0.014    & 323.0     & 382.0     \\
sig\_n    & 0.034  & 0.002 & 0.03     & 0.038     & 0.0        & 0.0      & 562.0     & 423.0    
\end{tabular}
\end{table}

\subsection{Concrete}

\begin{table}[H]
\centering
\begin{tabular}{l|l|l|l|l|l|l|l|l}
hyper     & mean  & sd    & hdi\_3\% & hdi\_97\% & mcse\_mean & mcse\_sd & ess\_bulk & ess\_tail \\ 
\hline
ls{[}0{]} & 3.667 & 0.537 & 2.86     & 4.776     & 0.059      & 0.042    & 80.0      & 77.0      \\ 
ls{[}1{]} & 5.278 & 0.741 & 4.125    & 6.865     & 0.093      & 0.066    & 65.0      & 117.0     \\ 
ls{[}2{]} & 5.558 & 1.264 & 3.415    & 7.863     & 0.135      & 0.095    & 100.0     & 64.0      \\ 
ls{[}3{]} & 2.933 & 0.497 & 2.19     & 3.976     & 0.054      & 0.041    & 104.0     & 98.0      \\ 
ls{[}4{]} & 3.757 & 0.636 & 2.897    & 4.969     & 0.069      & 0.049    & 81.0      & 77.0      \\ 
ls{[}5{]} & 8.633 & 1.716 & 5.898    & 11.525    & 0.148      & 0.112    & 142.0     & 38.0      \\ 
ls{[}6{]} & 4.453 & 0.624 & 3.324    & 5.633     & 0.065      & 0.047    & 95.0      & 78.0      \\ 
ls{[}7{]} & 1.037 & 0.085 & 0.877    & 1.2       & 0.012      & 0.008    & 51.0      & 78.0      \\ 
sig\_f    & 1.588 & 0.242 & 1.187    & 1.977     & 0.032      & 0.023    & 58.0      & 96.0      \\ 
sig\_n    & 0.307 & 0.009 & 0.293    & 0.323     & 0.001      & 0.001    & 99.0      & 52.0      \\ 
\end{tabular}
\end{table}

\subsection{Energy}

\begin{table}[H]
\centering
\begin{tabular}{l|l|l|l|l|l|l|l|l}
hyper     & mean   & sd    & hdi\_3\% & hdi\_97\% & mcse\_mean & mcse\_sd & ess\_bulk & ess\_tail \\ 
\hline
ls{[}0{]} & 2.788  & 1.231 & 1.215    & 5.005     & 0.114      & 0.081    & 93.0      & 117.0     \\ 
ls{[}1{]} & 3.738  & 1.886 & 1.459    & 7.848     & 0.233      & 0.172    & 89.0      & 102.0     \\ 
ls{[}2{]} & 0.887  & 0.073 & 0.763    & 1.04      & 0.006      & 0.005    & 128.0     & 44.0      \\ 
ls{[}3{]} & 2.92   & 1.186 & 1.209    & 5.078     & 0.131      & 0.093    & 73.0      & 78.0      \\ 
ls{[}4{]} & 2.892  & 1.426 & 1.056    & 5.868     & 0.153      & 0.108    & 100.0     & 67.0      \\ 
ls{[}5{]} & 25.615 & 3.875 & 19.263   & 32.822    & 0.367      & 0.26     & 99.0      & 75.0      \\ 
ls{[}6{]} & 1.93   & 0.165 & 1.668    & 2.261     & 0.021      & 0.015    & 65.0      & 78.0      \\ 
ls{[}7{]} & 21.52  & 2.68  & 16.378   & 26.616    & 0.228      & 0.164    & 139.0     & 78.0      \\ 
sig\_f    & 1.002  & 0.134 & 0.795    & 1.206     & 0.016      & 0.011    & 74.0      & 76.0      \\ 
sig\_n    & 0.045  & 0.001 & 0.043    & 0.048     & 0.0        & 0.0      & 86.0      & 93.0      \\ 
\end{tabular}
\end{table}
\newpage
\subsection{WineRed}

\begin{table}[H]
\centering
\begin{tabular}{l|l|l|l|l|l|l|l|l}
hyper      & mean  & sd    & hdi\_3\% & hdi\_97\% & mcse\_mean & mcse\_sd & ess\_bulk & ess\_tail  \\
\hline
ls{[}0{]}  & 2.867 & 0.803 & 1.637    & 4.057     & 0.085      & 0.064    & 135.0     & 77.0       \\
ls{[}1{]}  & 4.048 & 0.947 & 2.594    & 5.948     & 0.067      & 0.056    & 163.0     & 91.0       \\
ls{[}2{]}  & 4.101 & 1.291 & 2.474    & 6.917     & 0.104      & 0.078    & 170.0     & 77.0       \\
ls{[}3{]}  & 6.32  & 2.018 & 3.007    & 9.72      & 0.239      & 0.17     & 200.0     & 52.0       \\
ls{[}4{]}  & 3.806 & 1.135 & 1.815    & 5.552     & 0.096      & 0.073    & 143.0     & 77.0       \\
ls{[}5{]}  & 6.096 & 2.036 & 3.288    & 10.548    & 0.18       & 0.149    & 196.0     & 59.0       \\
ls{[}6{]}  & 3.99  & 1.029 & 2.415    & 6.372     & 0.137      & 0.097    & 67.0      & 65.0        \\
ls{[}7{]}  & 5.925 & 1.667 & 3.177    & 8.986     & 0.189      & 0.139    & 74.0      & 102.0       \\
ls{[}8{]}  & 4.065 & 1.405 & 1.894    & 6.46      & 0.166      & 0.132    & 109.0     & 55.0        \\
ls{[}9{]}  & 1.929 & 0.351 & 1.333    & 2.573     & 0.038      & 0.028    & 104.0     & 52.0       \\
ls{[}10{]} & 2.58  & 0.453 & 1.765    & 3.471     & 0.045      & 0.033    & 118.0     & 77.0        \\
sig\_f     & 0.698 & 0.095 & 0.547    & 0.875     & 0.013      & 0.01     & 76.0      & 34.0        \\
sig\_n     & 0.749 & 0.017 & 0.716    & 0.777     & 0.001      & 0.001    & 188.0     & 102.0      
\end{tabular}
\end{table}


\end{appendices}

\end{document}

%% file: tables/uci_regression.tex
\begin{table*}
\centering
\caption{A comparison of Sparse GP approaches for UCI benchmarks. RMSE ($\pm$ standard error of mean) evaluated on average of 10 splits with 80$\%$ of the data used for training. $\delta$ indicates that the posterior over hyperparameters is approximated by a point estimate under the respective scheme.}
\resizebox{\columnwidth}{!}{
\begin{tabular}{l|l|l|l|l|l|l|l|l}
\toprule
Dataset & $N$ & $d$ & GPR + HMC & SGPR & SGPR + HMC & SVGP & JointHMC  & FBGP\\
\midrule
$|M|$ & - & - & - & 100 & 100 & 100 & 100 & 100\\
\midrule
$q(\bm{\theta})$ & - & - & free-form & $\delta$ & free-form \textcolor{blue}{(ours)} & $\delta$ & free-form & free-form $(Z, \theta)$\\
\midrule
Boston & 506 & 13 & 3.049 (0.14) & 3.291 (0.11) & 3.286 (0.09) & 3.619 (0.11) & 3.28 (0.11) & 3.845 (0.103)\\
Concrete & 1030 & 8 & 4.864 (0.12) & 5.459 (0.09) & 5.402 (0.05) & 5.617 (0.09) &  5.612 (0.09) & 6.084 (0.11) \\
Energy & 768 & 8 & 0.441 (0.01) & 0.477 (0.008)& 0.469 (0.009) & 0.500 (0.01) & 0.755 (0.02) & 0.490 (0.011)\\
WineRed & 1599 & 11 & 0.640 (0.01) & 0.636 (0.008) & 0.635 (0.008) & 0.641 (0.007) &  0.641 (0.007) & 0.642 (0.007)\\
Yacht & 308 & 6 & 0.353 (0.03) & 0.412 (0.03) & 0.387 (0.03) & 0.606 (0.04)  & 0.794 (0.07) & 0.569 (0.037)\\
\bottomrule
\end{tabular}}
\end{table*}

%% file: tables/ablation_nlpds.tex
\begin{table}[H]
\centering
\caption{An ablation study for the doubly collapsed Sparse GPR scheme comparing performance with and without adapting the inducing locations during training. We report test NLPDs and RMSEs over 10 splits.}
\vspace{5pt}
\resizebox{\columnwidth}{!}{
\begin{tabular}{l|c|c|c|c|c|c}
\toprule
Dataset & Metric & Boston & Concrete & Energy & WineRed & Yacht \\
\midrule
Fixed Z & \multirow{2}{*}{RMSE} & 3.624 (0.110) & 6.021 (0.12) & 0.499 (0.014) & 0.640 (0.007) & 0.533 (0.036) \\
Adapt Z (\textcolor{blue}{ours})& & \textbf{3.286} (0.090) & \textbf{5.405} (0.07) & \textbf{0.469} (0.009) & 0.635 (0.008) & \textbf{0.387} (0.030)\\
\midrule
Fixed Z & \multirow{2}{*}{NLPD} & 2.640 (0.020) & 3.200 (0.06) & 0.696 (0.014) & 0.969 (0.012) & 0.791 (0.130) \\
Adapt Z (\textcolor{blue}{ours}) &  &\textbf{2.524} (0.022) & \textbf{3.065} (0.01) & \textbf{0.644} (0.013) & 0.961 (0.011) & \textbf{0.391} (0.146) \\
\bottomrule
\end{tabular}}
\label{tab:anlpds}
\end{table}